\documentclass[letterpaper]{article} 
\usepackage{aaai25}  
\usepackage{times}  
\usepackage{helvet}  
\usepackage{courier}  
\usepackage[hyphens]{url}  
\usepackage{graphicx} 
\usepackage{natbib}  
\usepackage{caption} 
\usepackage{algorithm}
\usepackage{amsmath,amssymb,amsfonts}
\usepackage{algorithmic}
\usepackage{textcomp}
\usepackage{xcolor}
\usepackage{enumitem}
\usepackage{todonotes}
\usepackage{tabularx}
\usepackage{booktabs}
\usepackage{comment}
\usepackage{svg}
\def\BibTeX{{\rm B\kern-.05em{\sc i\kern-.025em b}\kern-.08em
    T\kern-.1667em\lower.7ex\hbox{E}\kern-.125emX}}

\usepackage{newfloat}
\usepackage{listings}
\DeclareCaptionStyle{ruled}{labelfont=normalfont,labelsep=colon,strut=off} 
\floatstyle{ruled}
\newfloat{listing}{tb}{lst}{}
\floatname{listing}{Listing}
\title{Prior knowledge Injection into Deep Learning Models Predicting Gene Expression from Whole Slide Images}
\author{
    Max Hallemeesch\textsuperscript{\rm 1}\equalcontrib,
    Marija Pizurica\textsuperscript{\rm 1, \rm 2}\equalcontrib,
    Paloma Rabaey\textsuperscript{\rm 1},
    Olivier Gevaert\textsuperscript{\rm 2}\textdagger,
    Thomas Demeester\textsuperscript{\rm 1}\textdagger,
    Kathleen Marchal\textsuperscript{\rm 1}\textdagger
}
\affiliations{
    \textsuperscript{\rm 1}Internet Technology
and Data Science Lab (IDLab), Ghent University -- imec, Ghent, Belgium\\
    \textsuperscript{\rm 2}Department of Medicine, Stanford Center for Biomedical Informatics Research (BMIR), Stanford University, Stanford, USA\\
    max@hallemeesch.net, marija.pizurica@ugent.be, paloma.rabaey@ugent.be, ogevaert@stanford.edu, thomas.demeester@ugent.be, kathleen.marchal@ugent.be, 

}

\begin{document}

\maketitle

\begin{abstract}
Cancer diagnosis and prognosis primarily depend on clinical parameters such as age and tumor grade, and are increasingly complemented by molecular data, such as gene expression, from tumor sequencing. However, sequencing is costly and delays oncology workflows. Recent advances in Deep Learning allow to predict molecular information from morphological features within Whole Slide Images (WSIs), offering a cost-effective proxy of the molecular markers. While promising, current methods lack the robustness to fully replace direct sequencing. Here we aim to improve existing methods by introducing a model-agnostic framework that allows to inject prior knowledge on gene-gene interactions into Deep Learning architectures, thereby increasing accuracy and robustness. We design the framework to be generic and flexibly adaptable to a wide range of architectures.  
In a case study on breast cancer, our strategy leads to an average increase of 983 significant genes (out of 25,761) across all 18 experiments, with 14 generalizing to an increase on an independent dataset. 
Our findings reveal a high potential for injection of prior knowledge to increase gene expression prediction performance from WSIs across a wide range of architectures.
\end{abstract}

\begin{links}
    \link{Code}{https://github.com/MaxHallemeesch/PRALINE}
\end{links}

\section{Introduction}

In clinical settings, cancer diagnosis and prognosis predominantly rely on clinical parameters such as age, tumor grade, and histopathological evaluation of tissue sections by pathologists. However, cancer cells form a complex ecosystem, exhibiting intricate molecular profiles and cellular evolution that result in substantial intra- and inter-patient heterogeneity \cite{hausser2020tumour}. This complexity poses significant challenges to assess cancer aggressiveness and determine the most effective therapeutic strategies.

To unravel the underlying processes driving aggressive cancer outcomes in a more targeted way, tissue samples are increasingly subjected to RNA sequencing (RNA-Seq) to extract gene expression profiles. This progress has significantly enhanced our understanding of cancer heterogeneity and has led to the identification of genes whose expression is linked to prognosis and treatment sensitivity \cite{vasaikar2019proteogenomic, chawla2022gene}. However, incorporating RNA-Seq into oncology workflows requires time-consuming and expensive laboratory procedures, limiting their integration into routine practice. 

Recent advances in the field of Deep Learning and Computer Vision allow for predicting molecular information from scanned histopathology tissue sections, also referred to as Whole Slide Images (WSIs), which are abundantly available as they are obtained in standard clinical practice \cite{sequoia, relatedwork}. These methods offer potential to drastically reduce costs and delays in current diagnostic and treatment workflows, and in research settings to provide new insights on how molecular characteristics drive tumor aggressiveness.

\begin{figure}
    \begin{center}
    \includegraphics[width=0.4\textwidth]{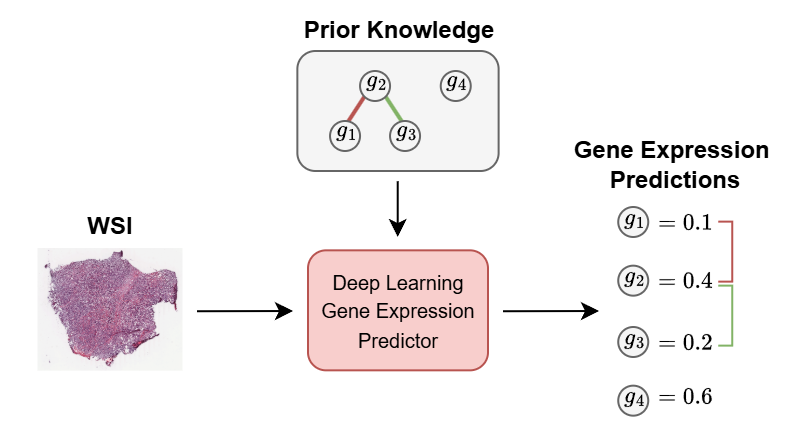}
    \end{center}
    \caption{High level overview. We inject prior knowledge, here indicating that genes $g_1$ and $g_3$ are strongly correlated with $g_2$, into the gene expression Predictor. This guides the Predictor towards extracting correlated gene expression predictions from the WSI (e.g., for high $g_2$ expression, the expressions of $g_1$ and $g_3$ are encouraged to also be high). 
    }
    \label{fig:main drawing}
\end{figure}

Current models predicting gene expression from WSIs \cite{alsaafin2023learning, schmauch2020deep, sequoia} have demonstrated potential, but are not yet robust enough to replace RNA-Seq profiling of the tissue. Here, we present a generic and flexible framework to inject prior knowledge on gene-gene interactions into current gene expression predictors (Figure \ref{fig:main drawing}). By directly inserting the prior knowledge into the model, as opposed to expecting the model to infer complex gene-gene relations indirectly, we guide the model towards realistic gene expression predictions and thereby increase prediction robustness. The main contributions of our research are threefold:

\begin{enumerate}
    \item We present a method to transform prior gene-gene interaction knowledge into gene embeddings, serving as abstract representations of relational gene properties.
    \item We present a simple model-agnostic framework to inject these gene embeddings into Deep Learning models predicting gene expression from WSIs, to improve their accuracy and robustness.
    \item To show its generality, we apply our method to several model architectures, and evaluate how it increases prediction accuracy, in the development dataset (TCGA-BRCA) and an independent dataset (CPTAC-BRCA). 
\end{enumerate}

\section{Related Work}
We first cover the overall workflow of WSI-based gene expression prediction models. Next, we summarize related work for injecting gene co-expression prior knowledge into these models, and we highlight how our method differs.

\subsection{Gene Expression Prediction from WSIs} \label{sec:gene_expression_prediction}

\begin{figure*}
    \begin{center}
    \includegraphics[scale=0.25]{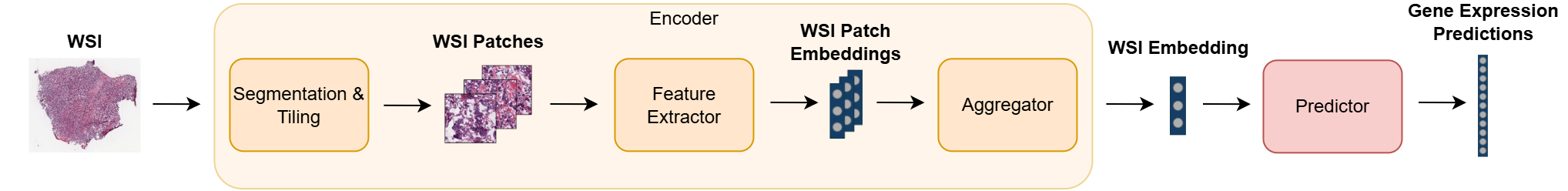}
    \end{center}
    \caption{General workflow. First, the WSI is processed by an Encoder, which extracts patches, their corresponding features, and aggregates them into a single WSI embedding. Then, a Predictor transforms the embedding into gene expression predictions. }
    \label{fig:pipeline}
\end{figure*}

Multiple architectures exist to predict gene expression from WSIs. These models usually follow the same workflow, yet consist of different subcomponents (Figure \ref{fig:pipeline}). First, WSIs are pre-processed and tiled into square patches. Next, $N_{patches}$ patches are randomly sampled and patch embeddings are generated using a feature extractor. An embedding $\textbf{w} \in \mathbb{R}^{1 \times d}$ to represent the WSI is computed by aggregating the patch embeddings. Then, this WSI embedding is passed to a Predictor consisting of a single linear layer, which finally produces the gene expression predictions. For conciseness, we will also refer to the components that transform the WSI into a WSI embedding as the Encoder (Segmentation \& Tiling, Feature Extractor, Aggregator from Figure 2).

In WSI literature, different implementations exist for the Feature Extractor and Aggregator components. To demonstrate the applicability of our approach regardless of choices made for these components, we implement two versions of commonly used feature extractors and three variations for patch embedding aggregators. We use foundation models tailored specifically for WSI feature extraction, such as CTransPath \cite{wang2022transformer} and UNI \cite{unifeatures}. For patch aggregation, we consider three alternatives in recent WSI literature for gene expression prediction: a regular MLP, as used in HE2RNA \cite{schmauch2020deep}, a transformer with self-attention as employed by tRNAsformer \cite{alsaafin2023learning} and finally a transformer linearized with SummaryMixing \cite{summarymixing}, as implemented in SEQUOIA \cite{sequoia}.

While each of these architectures can associate morphological features within WSIs to gene expression levels to a certain extent, their predictions lack robustness. Here, we hypothesize that the gene predictions could be made more reliable by injecting the model with prior knowledge on gene-gene interactions (Figure \ref{fig:main drawing}).

\subsection{Prior knowledge exploitation}
While there has been substantial previous work on gene expression prediction from WSIs, there is limited research on how to integrate prior knowledge on gene-gene interactions into these models. 

\citet{relatedwork} propose using gene-gene interaction prior knowledge to cluster genes into $K=50$ groups of related genes. Then, they train $K$ independent models to predict gene expression levels per cluster. They achieved optimal results by using gene co-expression profiles as prior knowledge, suggesting that co-expressed genes share morphological features in WSIs, enabling better generalization within clusters.

Instead of opting for multiple models trained in parallel, we prefer the use of a single prior knowledge-enhanced model that predicts the entire transcriptome jointly. This approach not only enables more efficient utilization of the model architecture and training resources -- training a single model instead of $K$ separate ones -- but also enhances the integration of prior knowledge directly into the model architecture rather than relying on it as a pre-processing step.

\section{Methodology}
\begin{figure}
    \begin{center}
    \includegraphics[width=0.5\textwidth]{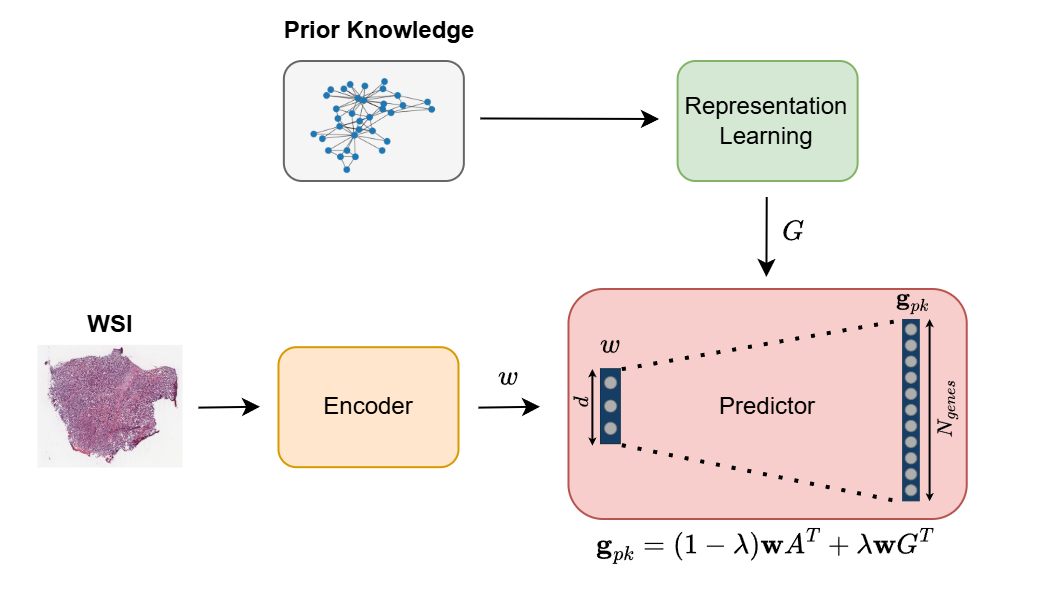}
    \end{center}
    \caption{Overview of our framework. First, we transform prior knowledge (gene-gene interaction network) into gene embeddings $G$ using a representation learning technique. The Encoder transforms the WSI into an embedding $\textbf{w}$. We then inject the gene embeddings into the Predictor by linearly transforming $\textbf{w}$ into gene predictions $\mathbf{g_{pk}}$ using a weighted sum of the linear predictor layer $A$, and the gene embeddings $G$. Hyperparameter $\lambda$ controls the effect of prior knowledge.
    }
    \label{fig:framework}
\end{figure}

We propose a generic framework for integrating prior knowledge on gene-gene interactions into gene expression prediction models from WSIs. The injection of prior knowledge occurs within the Predictor, making it the only non-interchangeable component of the framework (Figure \ref{fig:framework}).

\begin{table}
    \centering
    \resizebox{0.45\textwidth}{!}{
        \begin{tabular}{cccc} \toprule
             & \textbf{External} & \textbf{Internal} & \textbf{Combined}\\\midrule
             Number of Genes in Network  & 4 646 & 1 952 & 6 021\\
            Co-expressed Pairs of Genes & 41 672 & 48 858 & 88 402\\
            \bottomrule
        \end{tabular}
    }
        \caption{Overview of the three gene-gene interaction networks that serve as prior knowledge in our approach. For the internal and combined prior knowledge, numbers are averaged out over five folds. We only consider genes that are present in the TCGA-BRCA expression annotations.}
    \label{tab:priorknowledge summary}
\end{table}

\subsection{Sources of prior knowledge}
The prior knowledge we exploit is represented by an undirected and unweighted graph in which nodes correspond with genes, and edges represent the co-expression property. We define a pair of genes ($i$, $j$) to be co-expressed if the Pearson correlation $R_{ij}$ between their respective expression levels ($p_i$, $p_j$) exceeds a threshold $\tau$. Determining a single ground truth gene co-expression graph is infeasible. Various factors, such as the variability of patient tumor properties across datasets, can contribute to the formation of different gene networks. We therefore consider three distinct sources of prior knowledge. Table \ref{tab:priorknowledge summary} shows an overview of the properties of these three sources of prior knowledge. 

First, we collected \textbf{external prior knowledge} from the Cancer Gene Neighborhood (CGN) subcollection of the Computational Gene Sets of the Human Molecular Signatures (MsigDB) dataset \cite{cgn}. This prior knowledge is independent of the dataset we use for model development (TCGA-BRCA), since it is extracted from an external source reflecting general gene-gene co-expression relationships that are not exclusively observed in breast cancer tissue, but are present in normal and carcinoma tissue of various cancer subtypes. This source of prior knowledge may therefore contribute towards an increased generalization performance. The co-expression network is extracted using a threshold of $\tau = 0.85$.

We also created an \textbf{internal prior knowledge} network that is extracted from TCGA-BRCA by hard thresholding the pairwise Pearson correlation for each pair of genes. We employ the same threshold $\tau=0.85$ as before. Among the pairs of genes co-expressed in the external prior knowledge, only a small subset was also co-expressed in the TCGA-BRCA dataset. We therefore constructed a \textbf{combined prior knowledge} set by taking the union of both sources.

\subsection{Prior knowledge representation learning}
The three prior knowledge co-expression graphs gathered in the previous section can each be represented by an adjacency matrix. However, injecting the prior knowledge into the prediction models in raw matrix format is infeasible, due to its high dimension and sparsity. Therefore, we compress this high-dimensional information into a low-dimensional representation that retains only the most meaningful properties of the original prior knowledge network. To this end, we employ a Matrix Factorization (MF) technique on the adjacency matrix to learn a prior knowledge embedding for each gene.

We opt for Nonnegative Matrix Factorization (NMF) due to its extensive usage in the field of Computational Biology \cite{NMFsource1, NMFsource2, NMFsource3}. NMF decomposes the sparse adjacency matrix $M\in \mathbb{R}^{N_{genes} \times N_{genes}}$ into two components $G, Y\in \mathbb{R}^{N_{genes} \times d}$ by minimizing the Frobenius Norm (see Equation (\ref{Frobenius})), such that the matrix product $GY^T$ is a reconstructed version of the input. In this notation, $Y$ represents a collection of basis elements in the $d$-dimensional latent space. Each row of $G$ can then be interpreted as an embedding representation of size $d$ for the corresponding gene as a linear combination of the basis elements in $Y$. Appendix \ref{app:embedding_quality} assesses the quality of these embeddings by determining how well the co-expression property is retained in latent space. 

\begin{equation}\label{Frobenius}
    \lVert M - GY^T\rVert^2_F = \sqrt{\sum_{(i, j)} |(M - GY^T)_{ij}|^2}
\end{equation}

\subsection{Framework Architecture}
We opt to inject the prior knowledge (PK) directly into the component responsible for making the gene expression prediction (Predictor; see Figure \ref{fig:pipeline}), adapting only the final linear layer $f_{linear}$ for simplicity (Figure \ref{fig:framework}). Our modifications result in the new layer $f_{pk}$ which transforms the WSI embedding $\textbf{w}$ into PK-informed predictions $\textbf{g}_{pk}$ (Equation \ref{eq:f_pk}). 

Hereto, we augment $f_{linear} (= \textbf{w}A^T + \textbf{b})$ with a new term, $\textbf{w}G^T$, where the i-th row of the embedding matrix $G$ represents the embedding for gene $i$. Hyperparameter $\lambda \in [0, 1]$ controls its influence on the expression prediction (Equation \ref{eq:elaborate}). Genes without an embedding correspond with a zero row in $G$, ensuring the calculation defaults to $f_{linear}$. As a result, prior knowledge is leveraged only for genes included in the gene co-expression network, preserving standard operations for all the others. To this end, our framework is broadly applicable to various models, provided the final predictor layer can be replaced with our proposed design. 

\begin{equation}\label{eq:f_pk}
    f_{pk} : \textbf{w} \xrightarrow{} \textbf{g}_{pk}
\end{equation}

\begin{equation}\label{eq:elaborate}
\begin{gathered}
    \textbf{g}_{pk} = (1-\lambda)\textbf{w}A^T + \lambda \textbf{w}G^T + \textbf{b} \\
    \textbf{w} \in \mathbb{R}^{1 \times d}, \textbf{g}_{pk} \in \mathbb{R}_{\geq 0}^{1 \times N_{genes}} \\
    A, G \in \mathbb{R}^{N_{genes} \times d},  \textbf{b} \in \mathbb{R}^{1 \times N_{genes}} 
\end{gathered}
\end{equation}
\vspace{1mm}

Consider two genes $i$ and $j$, with similar expression profiles, translating into similar embeddings $G_i$ and $G_j$. Consequently, $\textbf{w} G^T_i$ and $\textbf{w} G^T_j$ have a similar value, therefore guiding the predicted expression for these genes into the same direction. The new Predictor is depicted in red in Figure \ref{fig:framework}.

\begin{table*}
    \centering
    \resizebox{\textwidth}{!}{
    \begin{tabular}{lllllllllll} 
        \toprule
        \textbf{TCGA} & \textbf{No PK} & \textbf{External} & \textbf{Internal} & \textbf{Combined} & & \textbf{CPTAC} & \textbf{No PK} & \textbf{External} & \textbf{Internal} & \textbf{Combined} \\
        \cline{1-5} \cline{7-11} \noalign{\vspace{0.3ex}}
        \raggedright ctr\_mlp & $21,233$ & $22,225$ $(0.9)$ $\textcolor{teal}{{\scriptstyle \uparrow 992}}$ & $22, 278$ $(0.9)$ $\textcolor{teal}{{\scriptstyle \uparrow 1,045}}$& $22, 160$ $(0.9)$ $\textcolor{teal}{{\scriptstyle \uparrow 927}}$ & & \raggedright ctr\_mlp & $16,936$ & $16,313$ $\textcolor{purple}{{\scriptstyle \downarrow 623}}$& $\mathbf{18,116}$ $\textcolor{teal}{{\scriptstyle \uparrow 1,180}}$ & $\mathbf{17,363}$ $\textcolor{teal}{{\scriptstyle \uparrow 427}}$\\
        \raggedright ctr\_tf & $19,155$ & $21,647$ $(0.9)$ $\textcolor{teal}{{\scriptstyle \uparrow 2,429}}$ & $20,618$ $(0.2)$ $\textcolor{teal}{{\scriptstyle \uparrow 1,463}}$ & $20,548$ $(0.8)$ $\textcolor{teal}{{\scriptstyle \uparrow 1,393}}$ & & \raggedright ctr\_tf & $15,677$ & $15,146$ $\textcolor{purple}{{\scriptstyle \downarrow 531}}$ & $14,983$ $\textcolor{purple}{{\scriptstyle \downarrow 694}}$ & $14,386$ $\textcolor{purple}{{\scriptstyle \downarrow 1,291}}$ \\
        \raggedright ctr\_smx & $20,945$ & $22,564$ $(0.9)$ $\textcolor{teal}{{\scriptstyle \uparrow 1,619}}$ & $21,451$ $(0.5)$ $\textcolor{teal}{{\scriptstyle \uparrow 506}}$ & $21,944$ $(0.5)$ $\textcolor{teal}{{\scriptstyle \uparrow 999}}$ & & \raggedright ctr\_smx & $15,714$ & $16,682$ $\textcolor{teal}{{\scriptstyle \uparrow 968}}$ & $15,731$ $\textcolor{teal}{{\scriptstyle \uparrow 17}}$ & $15,753$ $\textcolor{teal}{{\scriptstyle \uparrow 39}}$  \\
        \raggedright uni\_mlp & $21,721$ & $22,666$ $(0.8)$ $\textcolor{teal}{{\scriptstyle \uparrow 945}}$ & $22,802$ $(0.8)$ $\textcolor{teal}{{\scriptstyle \uparrow 1,081}}$ & $\mathbf{23,214}$ $(0.9)$ $\textcolor{teal}{{\scriptstyle \uparrow 1,493}}$ & & \raggedright uni\_mlp & $15,560$ & $16,106$ $\textcolor{teal}{{\scriptstyle \uparrow 546}}$ & $16,045$ $\textcolor{teal}{{\scriptstyle \uparrow 485}}$ & $15,784$  $\textcolor{teal}{{\scriptstyle \uparrow 224}}$ \\
        \raggedright uni\_tf & $22, 124$ & $22, 461$ $(0.2)$ $\textcolor{teal}{{\scriptstyle \uparrow 337}}$& $22, 645$ $(0.1)$ $\textcolor{teal}{{\scriptstyle \uparrow 521}}$&  $22, 597$ $(0.1)$ $\textcolor{teal}{{\scriptstyle \uparrow 473}}$ & & \raggedright uni\_tf & $14,705$ & $15,648$ $\textcolor{teal}{{\scriptstyle \uparrow 763}}$& $15,469$ $\textcolor{teal}{{\scriptstyle \uparrow 764}}$& $15,400$ $\textcolor{teal}{{\scriptstyle \uparrow 695}}$\\
        \raggedright uni\_smx & $\mathbf{22,997}$ & $\mathbf{23,578}$ $(0.5)$ $\textcolor{teal}{{\scriptstyle \uparrow 581}}$ & $\mathbf{23,732}$ $(0.1)$ $\textcolor{teal}{{\scriptstyle \uparrow 735}}$ & $\mathbf{23,162}$ $(0.9)$ $\textcolor{teal}{{\scriptstyle \uparrow 165}}$ & &\raggedright uni\_smx & $16,952$ & $\mathbf{17,280}$ $\textcolor{teal}{{\scriptstyle \uparrow 328}}$ & $\mathbf{17,091}$  $\textcolor{teal}{{\scriptstyle \uparrow 139}}$ & $\mathbf{16,981}$ $\textcolor{teal}{{\scriptstyle \uparrow 29}}$  \\
        \bottomrule
    \end{tabular}
    }
    \caption{Number of significant genes on TCGA-BRCA (left) and CPTAC-BRCA (right) datasets, using no prior knowledge (No PK), or PK from External, Internal or Combined data sources. The number between parentheses shows optimal $\lambda$ value from Equation \ref{eq:elaborate}. ctr: CTrans feature extractor, uni: UNI feature extractor, mlp: Multilayer Perceptron, tf: Transformer, smx: SummaryMixing. The arrow up/down shows the difference with No PK. The top 5 models are indicated in bold in each table.}
    \label{eval tcga-cptac-brca}
\end{table*}

\subsection{Implementation details}

We consider two feature extractors (CTrans and UNI), and three aggregators (MLP, Transformer and SummaryMixing). This results in six model architectures, each trained separately and evaluated across: no prior knowledge (PK), external, internal and combined PK. All models are trained using the same training paradigm as in SEQUOIA \cite{sequoia}. In this approach, the MSE loss is used for training, while the Pearson correlation between the predicted and ground truth gene expression values are also considered for early stopping. We train each model configuration with $\lambda \in [0.1, 0.2, 0.5, 0.8, 0.9]$, and select optimal $\lambda$ in terms of number of significant genes. Genes are deemed significantly well-predicted if their predicted expression correlates well with the ground truth expression, which is quantified through a set of statistical conditions outlined in SEQUOIA \cite{sequoia}.

Data generated by the TCGA Research Network (https://www.cancer.gov/tcga) is used for training, consisting of 1133 WSIs of breast adenocarcinoma annotated with ground truth bulk RNA-Seq gene expression. We employ 5-fold cross validation, dividing the WSIs into patient-level splits. Each fold consists of a training (72\%), validation (8\%) and testing (20\%) subset. The optimal stopping point for training is determined on the validation set. To avoid leakage, the prior knowledge for each fold is only determined from the training and validation subset.

To assess generalization to an independent test set, we consider 106 WSIs of breast cancer from the Clinical Proteomic Tumor Analysis Consortium (NCI/NIH) (CPTAC-BRCA). Neither the external nor internal PK is derived from CPTAC-BRCA. For each model configuration, we employ the optimal $\lambda$ from the TCGA-BRCA test set.

\section{Results and Discussion}
We evaluated the number of significantly predicted genes across 18 experiments, including three sources of prior knowledge and six deep learning architectures, on both TCGA-BRCA and CPTAC-BRCA (Table \ref{eval tcga-cptac-brca}). Across 14 experiments, we observed an increase in the number of significant genes on both TCGA and CPTAC, demonstrating an enhanced generalization performance. 

On TCGA, all sources of prior knowledge (PK) increased the number of significantly predicted genes, across all six model architectures. The External source led to the highest increase of 1,150 on average, followed by the Combined with 908 genes, and the Internal source with 891 genes. Across different PK and aggregation schemes, increases for CTrans features were higher than for UNI. The highest overall increase was found in CTrans features with transformer aggregation. Even though the PK was more effective in the CTrans models, the UNI models consistently reached higher predicted genes than their CTrans counterparts, confirming UNI to be a superior feature extractor, as found in other WSI tasks \cite{unifeatures}. The overall best-performing model was UNI with SummaryMixing and Internal PK, (23,732 significantly predicted genes), closely followed by the same model with External PK (23,578 genes).

Although highly effective on TCGA, all PK sources with CTrans features showed bad generalization in CPTAC, with a decrease in number of genes in 4 out of 9 combinations (3 aggregators, 3 PK sources), indicating the effect of PK with CTrans features to be unreliable. With UNI features, all PK sources did result in an increase of predicted genes across all model combinations, rendering them superior to CTrans features across all PK methods for all aggregators except MLP.
The (somewhat surprising) large added value of PK in CTrans with MLP led to the best performing model combination on CPTAC with Internal PK (18,116 well-predicted genes), closely followed by Combined PK (17,363). The best-performing model on TCGA (UNI with SummaryMixing) also achieved competitive performance on CPTAC (17,280 genes with External PK).

In summary, PK helps to consistently increase the number of significantly predicted genes for UNI features. Across both TCGA and CPTAC, UNI with SummaryMixing and External PK appears the most reliable model combination for high performance while maintaining robustness.

\section{Conclusion}
We present a model agnostic framework to inject prior knowledge (PK) on gene-gene interactions into deep learning models predicting gene expression from WSIs. In our research, we consider three different sources of PK, two feature extractors and three aggregators. On TCGA-BRCA, our PK injection method led to an increase of 983 genes on average (across all 18 models), which transferred to an increase in the CPTAC-BRCA dataset in 14 out of 18 cases. We conclude that injecting PK has potential to improve gene prediction performance and robustness across a wide range of architectures.

\section{Limitations \& Future Work}
In our research, we only considered WSIs of breast cancer tissue as a case study to evaluate our framework on. However, prior knowledge is also available for other cancer types, underscoring the potential to extend this research beyond breast cancer. A limitation of the current work is the absence of an investigation into the quality of the prior knowledge and its subsequent impact on the downstream task. Interesting insights could be collected through an ablation study on prior knowledge.

\section{Acknowledgments}

Marija Pizurica's research is funded by the Research Foundation Flanders (FWO Vlaanderen) with grant numbers 116223N and V467423N. Paloma Rabaey’s research is also funded by the Research Foundation Flanders, with grant number 1170124N. 

\vspace{12pt}


\begin{thebibliography}{14}
\providecommand{\natexlab}[1]{#1}

\bibitem[{Alsaafin et~al.(2023)Alsaafin, Safarpoor, Sikaroudi, Hipp, and Tizhoosh}]{alsaafin2023learning}
Alsaafin, A.; Safarpoor, A.; Sikaroudi, M.; Hipp, J.~D.; and Tizhoosh, H. 2023.
\newblock Learning to predict RNA sequence expressions from whole slide images with applications for search and classification.
\newblock \emph{Communications Biology}, 6(1): 304.

\bibitem[{Chawla et~al.(2022)Chawla, Rockstroh, Lehman, Ratther, Jain, Anand, Gupta, Bhattacharya, Poonia, Rai et~al.}]{chawla2022gene}
Chawla, S.; Rockstroh, A.; Lehman, M.; Ratther, E.; Jain, A.; Anand, A.; Gupta, A.; Bhattacharya, N.; Poonia, S.; Rai, P.; et~al. 2022.
\newblock Gene expression based inference of cancer drug sensitivity.
\newblock \emph{Nature communications}, 13(1): 5680.

\bibitem[{Chen et~al.(2024)Chen, Ding, Lu, Williamson, Jaume, Chen, Zhang, Shao, Song, Shaban et~al.}]{unifeatures}
Chen, R.~J.; Ding, T.; Lu, M.~Y.; Williamson, D.~F.; Jaume, G.; Chen, B.; Zhang, A.; Shao, D.; Song, A.~H.; Shaban, M.; et~al. 2024.
\newblock Towards a General-Purpose Foundation Model for Computational Pathology.
\newblock \emph{Nature Medicine}.

\bibitem[{Devarajan(2008)}]{NMFsource1}
Devarajan, K. 2008.
\newblock Nonnegative matrix factorization: an analytical and interpretive tool in computational biology.
\newblock \emph{PLoS computational biology}, 4(7): e1000029.

\bibitem[{Hausser and Alon(2020)}]{hausser2020tumour}
Hausser, J.; and Alon, U. 2020.
\newblock Tumour heterogeneity and the evolutionary trade-offs of cancer.
\newblock \emph{Nature Reviews Cancer}, 20(4): 247--257.

\bibitem[{Liu et~al.(2017)Liu, Wang, Gao, Zheng, Xu, and Yu}]{NMFsource3}
Liu, J.-X.; Wang, D.; Gao, Y.-L.; Zheng, C.-H.; Xu, Y.; and Yu, J. 2017.
\newblock Regularized non-negative matrix factorization for identifying differentially expressed genes and clustering samples: A survey.
\newblock \emph{IEEE/ACM transactions on computational biology and bioinformatics}, 15(3): 974--987.

\bibitem[{Parcollet et~al.(2024)Parcollet, van Dalen, Zhang, and Bhattacharya}]{summarymixing}
Parcollet, T.; van Dalen, R.; Zhang, S.; and Bhattacharya, S. 2024.
\newblock SummaryMixing: A linearcomplexity alternative to self-attention for speech recognition and understanding.
\newblock \emph{arXiv preprint arXiv:2307.07421}.

\bibitem[{Pascual-Montano et~al.(2006)Pascual-Montano, Carmona-Saez, Chagoyen, Tirado, Carazo, and Pascual-Marqui}]{NMFsource2}
Pascual-Montano, A.; Carmona-Saez, P.; Chagoyen, M.; Tirado, F.; Carazo, J.~M.; and Pascual-Marqui, R.~D. 2006.
\newblock bioNMF: a versatile tool for non-negative matrix factorization in biology.
\newblock \emph{BMC bioinformatics}, 7: 1--9.

\bibitem[{Pizurica et~al.(2024)Pizurica, Zheng, Carrillo-Perez, Noor, Yao, Wohlfart, Vladimirova, Marchal, and Gevaert}]{sequoia}
Pizurica, M.; Zheng, Y.; Carrillo-Perez, F.; Noor, H.; Yao, W.; Wohlfart, C.; Vladimirova, A.; Marchal, K.; and Gevaert, O. 2024.
\newblock Digital profiling of gene expression from histology images with linearized attention.
\newblock \emph{Nature Communications}, 15(1): 9886.

\bibitem[{Schmauch et~al.(2020)Schmauch, Romagnoni, Pronier, Saillard, Maill{\'e}, Calderaro, Kamoun, Sefta, Toldo, Zaslavskiy et~al.}]{schmauch2020deep}
Schmauch, B.; Romagnoni, A.; Pronier, E.; Saillard, C.; Maill{\'e}, P.; Calderaro, J.; Kamoun, A.; Sefta, M.; Toldo, S.; Zaslavskiy, M.; et~al. 2020.
\newblock A deep learning model to predict RNA-Seq expression of tumours from whole slide images.
\newblock \emph{Nature communications}, 11(1): 3877.

\bibitem[{Subramanian et~al.(2005)Subramanian, Tamayo, Mootha, Mukherjee, Ebert, Gillette, Paulovich, Pomeroy, Golub, Lander et~al.}]{cgn}
Subramanian, A.; Tamayo, P.; Mootha, V.~K.; Mukherjee, S.; Ebert, B.~L.; Gillette, M.~A.; Paulovich, A.; Pomeroy, S.~L.; Golub, T.~R.; Lander, E.~S.; et~al. 2005.
\newblock Gene set enrichment analysis: a knowledge-based approach for interpreting genome-wide expression profiles.
\newblock \emph{Proceedings of the National Academy of Sciences}, 102(43): 15545--15550.

\bibitem[{Vasaikar et~al.(2019)Vasaikar, Huang, Wang, Petyuk, Savage, Wen, Dou, Zhang, Shi, Arshad et~al.}]{vasaikar2019proteogenomic}
Vasaikar, S.; Huang, C.; Wang, X.; Petyuk, V.~A.; Savage, S.~R.; Wen, B.; Dou, Y.; Zhang, Y.; Shi, Z.; Arshad, O.~A.; et~al. 2019.
\newblock Proteogenomic analysis of human colon cancer reveals new therapeutic opportunities.
\newblock \emph{Cell}, 177(4): 1035--1049.

\bibitem[{Wang et~al.(2022)Wang, Yang, Zhang, Wang, Zhang, Yang, Huang, and Han}]{wang2022transformer}
Wang, X.; Yang, S.; Zhang, J.; Wang, M.; Zhang, J.; Yang, W.; Huang, J.; and Han, X. 2022.
\newblock Transformer-based unsupervised contrastive learning for histopathological image classification.
\newblock \emph{Medical image analysis}, 81: 102559.

\bibitem[{Weitz et~al.(2022)Weitz, Wang, Kartasalo, Egevad, Lindberg, Gr{\"o}nberg, Eklund, and Rantalainen}]{relatedwork}
Weitz, P.; Wang, Y.; Kartasalo, K.; Egevad, L.; Lindberg, J.; Gr{\"o}nberg, H.; Eklund, M.; and Rantalainen, M. 2022.
\newblock Transcriptome-wide prediction of prostate cancer gene expression from histopathology images using co-expression-based convolutional neural networks.
\newblock \emph{Bioinformatics}, 38(13): 3462--3469.

\end{thebibliography}

\appendix
\section*{Appendix}

\section{Prior Knowledge Embedding Quality} \label{app:embedding_quality}
We assess embedding quality by determining how well the co-expression property is retained in latent space, which is quantified through Neighborhood Preservation, as shown in Equation (\ref{NeighborhoodPreservation}). 
The metric computes the average overlap in the top $k$ neighbors in both spaces, across all considered genes. The set of $k$ closest neighbors for gene $i$ in the high-dimensional and low-dimensional space are respectively represented by $HDN^k_i$ and $LDN^k_i$. For $k =$ 100, our external, internal, combined prior knowledge embeddings respectively preserve 34\%, 16\% and 35\% of their original neighbors.

\begin{equation}\label{NeighborhoodPreservation}
    NP(k) = \frac{1}{N_{genes}} \sum_{i} \frac{1}{k}| HDN^k_i \cap LDN^k_i | 
\end{equation}


\bibliographystyle{unsrt}

\end{document}